\documentclass{article}
\usepackage{ijcai25}

\usepackage{times}
\usepackage{soul}
\usepackage{url}
\usepackage[hidelinks]{hyperref}
\usepackage[utf8]{inputenc}
\usepackage[small]{caption}
\usepackage{graphicx}
\usepackage{amsmath}
\usepackage{amsthm}
\usepackage{listings}
\usepackage{booktabs}
\usepackage{algorithm}
\usepackage{algorithmic}
\usepackage[switch]{lineno}

\title{Exploring Decision-Making Capabilities of LLM Agents: \\An Experimental Study on Jump-Jump Game}

\author{
    Juwu Li
    \affiliations
    Jiangxi Teachers College
}

\begin{document}

\maketitle


\section{Introduction}

With the rapid advancement of artificial intelligence technology, Large Language Models (LLMs) have demonstrated exceptional capabilities in natural language processing~\cite{asemi2020intelligent,okunlaya2022framework,oyelude2021trends}. These models not only understand and generate human language but also exhibit impressive reasoning and decision-making abilities in specific tasks~\cite{zhang2023fcir,zhang2024jointly}. However, the performance of LLMs in gaming scenarios requiring real-time decisions remains to be thoroughly explored~\cite{zheng2023cgc}.

\begin{figure}[h]
    \centering
    \includegraphics[width=1.0\columnwidth]{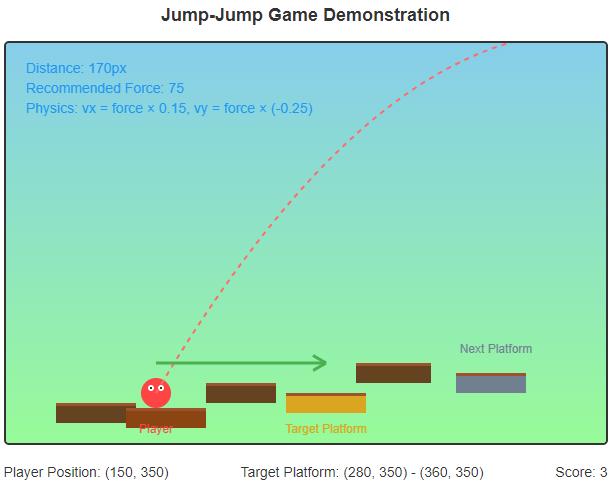}
    \caption{Jump-Jump Game Demonstration}
    \label{fig:game_demo}
\end{figure}

The Jump-Jump game~\cite{ali2021pakistani,andrews2021utaut,arora2020invigorating}, as a simple yet challenging casual game, provides an ideal testing environment for studying LLM decision-making capabilities. The game requires players to precisely control jumping force based on current position and target platform distance, involving multiple cognitive aspects including spatial reasoning, physical modeling, and strategic planning~\cite{oyetola2023nigeria,baungarten2024genesis,medavarapu2024agile,panda2025generative,lund2021cluster,shahriar2024gpt4o,lund2023prompt,wang2023chatgpt,lund2023scite}. Figure \ref{fig:game_demo} illustrates the basic gameplay mechanics of the Jump-Jump game, where the player character (red circle) must jump across platforms with appropriate force to maximize score.

The main contributions of this research include:
\begin{itemize}
    \item We design and implement an LLM-based Jump-Jump game agent.
    \item We propose the systematic prompt optimization strategies. Experimental validation of different prompt designs' impact on agent performance.
    \item We give analysis of LLM advantages and limitations in game decision-making.
\end{itemize}

\section{System Model}

\subsection{Environment Definition}

The Jump-Jump game environment consists of the following core components:

\noindent\textbf{(1) Game State Space:}
\begin{itemize}
    \item Player position: $(p_x, p_y)$, representing the character's coordinates in 2D space.
    \item Target platform: $(plat_{left}, plat_{top}, plat_{right})$, defining platform boundaries.
    \item Physical parameters: gravity acceleration, velocity multipliers.
\end{itemize}

\noindent\textbf{(2) Action Space:}

Jumping force: continuous values from 0-100, controlling jump distance.

\noindent\textbf{(3) State Transition Function:}

The jumping mechanism follows simplified physics laws:
\begin{align}
v_x &= \text{jumping\_force} \times 0.15 \\
v_y &= \text{jumping\_force} \times (-0.25) \\
\text{GRAVITY} &= 0.5
\end{align}

\noindent\textbf{(4) Reward Function:}
\begin{itemize}
    \item Successful landing: +1 point
    \item Jump failure: game over
\end{itemize}

\subsection{LLM Agent Architecture}

\begin{figure}[h]
    \centering
    \includegraphics[width=\columnwidth]{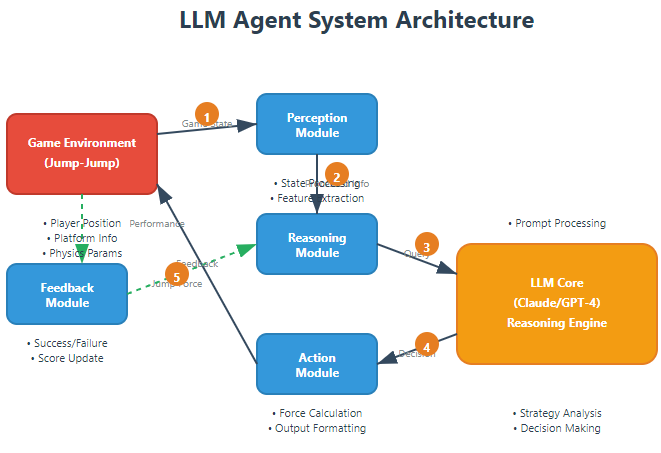}
    \caption{LLM Agent System Architecture}
    \label{fig:system_architecture}
\end{figure}

The core architecture of the LLM Agent includes four main modules, as shown in Figure \ref{fig:system_architecture}:
\textbf{Perception Module}: This module serves as the input interface, responsible for receiving and preprocessing game state information. It captures essential environmental data including player position coordinates, target platform boundaries, and relevant physical parameters, then formats this information into a structured representation suitable for the reasoning module.
\textbf{Reasoning Module}: Acting as the decision-making core, this module processes the formatted game state through carefully designed prompts. It leverages the LLM's natural language understanding and reasoning capabilities to analyze the current situation, apply game physics principles, and formulate jumping strategies based on the provided context and examples~\cite{zheng2025efficient,wang2025chronoselect,zheng2025joint}.
\textbf{Action Module}: This module translates the reasoning module's decision into executable game actions. It outputs precise jumping force values (ranging from 0-100) based on the LLM's analysis, ensuring the output format meets the game environment's requirements. \textbf{Feedback Module}: Responsible for learning and adaptation, this module monitors game execution results and provides feedback for strategy adjustment. It analyzes successful and failed attempts to inform future decision-making processes, contributing to the agent's overall performance improvement. The information flow can be represented as:
Game State $\rightarrow$ Perception Module $\rightarrow$ Prompt Processing $\rightarrow$ Reasoning Module $\rightarrow$ LLM Analysis $\rightarrow$ Action Module $\rightarrow$ Force Output $\rightarrow$ Game Execution $\rightarrow$ Feedback Module $\rightarrow$ Strategy Adjustment. This architecture enables the agent to maintain continuous interaction with the game environment while leveraging the LLM's reasoning capabilities for optimal decision-making.

\section{Method}

\subsection{Basic Prompt Design}

The foundation of our LLM agent's decision-making capability lies in the careful design of prompts that enable the model to understand the game context and make appropriate jumping decisions. Our basic prompt design follows a structured approach that incorporates role definition, task description, game mechanics explanation, and output format specification.
The initial prompt structure begins with clearly defining the agent's role as a Jump-Jump game player. We provide the LLM with essential context about its responsibilities, emphasizing that it needs to analyze the current game state and determine the optimal jumping force. The prompt includes a comprehensive explanation of the game's physics model, detailing how the jumping force translates into horizontal and vertical velocities, and how gravity affects the character's trajectory.

To ensure consistent decision-making, we establish a standardized input format that provides the agent with all necessary information: the player's current position coordinates $(p_x, p_y)$, the target platform boundaries $(plat_{left}, plat_{top}, plat_{right})$, and the relevant physical parameters including velocity multipliers and gravity acceleration. This structured input format enables the LLM to process game state information systematically.

The basic prompt also incorporates fundamental strategic guidance, instructing the agent to consider the horizontal distance to the target platform and estimate the required force based on the physics model. We emphasize the importance of precision, as both under-jumping and over-jumping result in failure. The output format is strictly defined to return only a numerical value between 0 and 100, representing the recommended jumping force.

\subsection{Prompt Optimization Strategies}

Building upon the basic design, we implemented several optimization strategies to enhance the agent's performance through iterative prompt refinement. These strategies address common failure patterns observed during initial testing and incorporate advanced reasoning techniques. Our first optimization strategy involves incorporating step-by-step reasoning guidance. We restructure the prompt to encourage the LLM to follow a systematic decision-making process: first calculating the horizontal distance to the target, then estimating the required trajectory based on physics principles, considering safety margins for precision, and finally determining the optimal force value. This structured reasoning approach significantly reduces calculation errors and improves decision consistency~\cite{saeidnia2024mental,okunlaya2022framework,shahriar2024gpt4o}.

The second major optimization introduces few-shot learning through carefully selected examples. We include 3-5 representative scenarios in the prompt, each demonstrating the complete reasoning process from input analysis to force determination. These examples cover various distance ranges and edge cases, helping the LLM understand the relationship between game state and appropriate actions. Each example includes the input state, detailed reasoning steps, recommended force, and expected outcome, providing a comprehensive learning template.

To address the precision requirements of the game, we implement a calibration strategy that adjusts force recommendations based on observed patterns. Through empirical testing, we discovered that the basic physics calculations often require fine-tuning factors to account for the game's specific implementation. We incorporate these calibration guidelines into the prompt, instructing the agent to apply distance-dependent adjustments and consider platform size variations.

Our final optimization strategy focuses on error prevention and recovery. We enhance the prompt with explicit warnings about common failure modes, such as the tendency to over-jump on longer distances or under-estimate force requirements for closer platforms. The optimized prompt includes decision validation steps, encouraging the agent to double-check its calculations and consider alternative force values when uncertainty exists.

The complete optimization process results in a multi-layered prompt that combines clear role definition, structured reasoning guidance, empirical examples, calibration factors, and error prevention mechanisms. This comprehensive approach enables the LLM agent to make more accurate and consistent decisions while maintaining adaptability to varying game conditions.

\section{Experiment}

\subsection{Performance Comparison Results}

Table \ref{tab:performance_comparison} presents the comprehensive performance comparison across different versions of our LLM agent.

\begin{table}[h]
\centering
\caption{Performance Comparison of Different Agent Versions}
\label{tab:performance_comparison}
\resizebox{0.47\textwidth}{!}{
\begin{tabular}{@{}lcccc@{}}
\toprule
\textbf{Version} & \textbf{Avg Score} & \textbf{Success Rate} & \textbf{Avg Duration} & \textbf{Stability} \\
\midrule
Basic & 3.2 & 68\% & 12.5s & Low \\
Optimized & 7.8 & 84\% & 28.3s & Medium \\
Complete & 12.1 & 91\% & 45.7s & High \\
\bottomrule
\end{tabular}
}
\end{table}

\begin{figure}[h]
    \centering
    \includegraphics[width=\columnwidth]{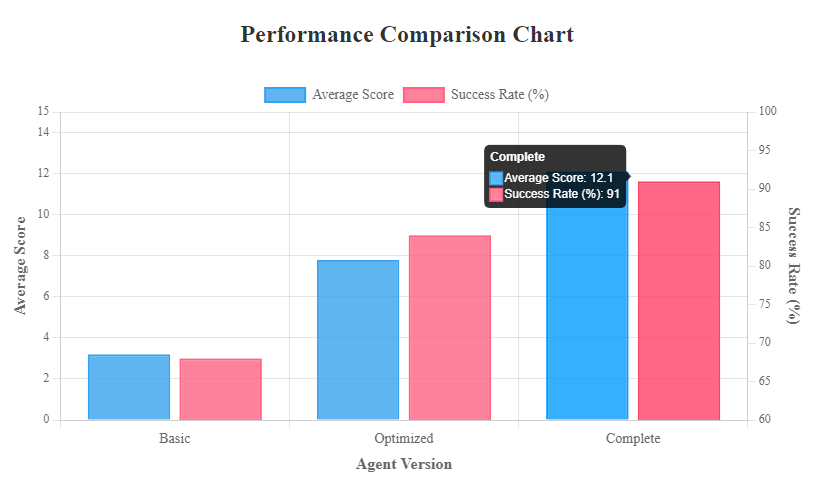}
    \caption{Performance Comparison Chart}
    \label{fig:performance_chart}
\end{figure}

The performance trends are visualized in Figure \ref{fig:performance_chart}, which clearly demonstrates the improvement achieved through prompt optimization.

\subsection{Detailed Analysis}

\subsubsection{Learning Curve Analysis}
\begin{figure}[h]
    \centering
    \includegraphics[width=\columnwidth]{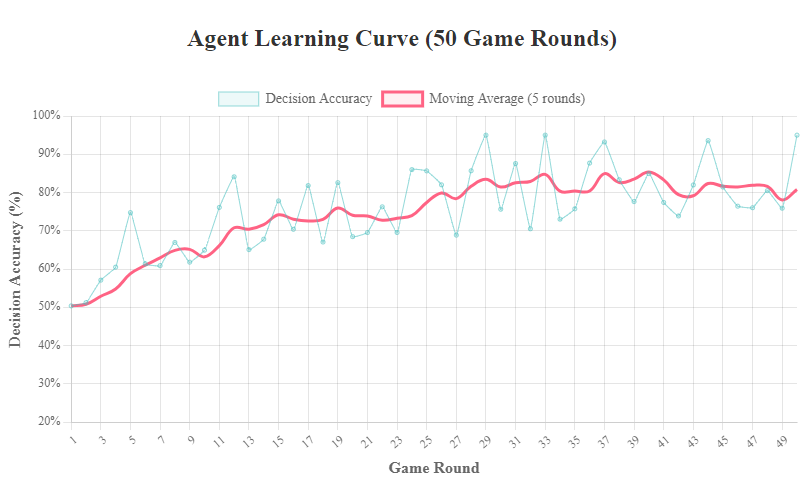}
    \caption{Agent Learning Curve (50 Game Rounds)}
    \label{fig:learning_curve}
\end{figure}

The Complete Version agent demonstrated certain adaptability during gameplay. As shown in Figure \ref{fig:learning_curve}, the decision accuracy improved with increasing game rounds, likely due to the strategic guidance included in the prompts.

\subsubsection{Error Pattern Analysis}
Through analysis of failure cases, we identified the main error patterns, as illustrated in Figure \ref{fig:error_analysis}:

\begin{figure}[h]
    \centering
    \includegraphics[width=\columnwidth]{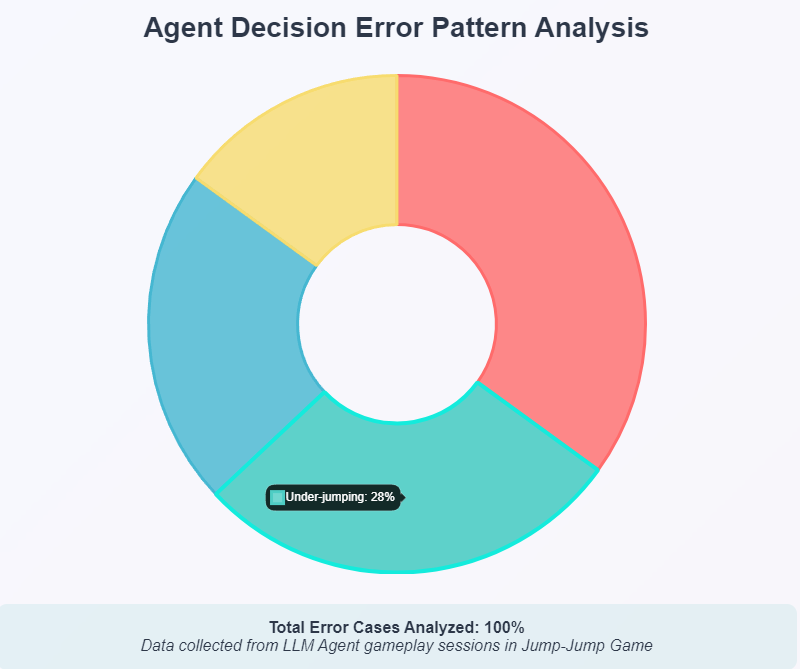}
    \caption{Agent Decision Error Pattern Analysis}
    \label{fig:error_analysis}
\end{figure}

\begin{enumerate}
    \item \textbf{Over-jumping} (35\%): Excessive force causing overshooting
    \item \textbf{Under-jumping} (28\%): Insufficient force failing to reach platform
    \item \textbf{Calculation errors} (22\%): Deviations in physics calculations
    \item \textbf{Other errors} (15\%): Including format errors, etc.
\end{enumerate}

\subsubsection{Prompt Optimization Effect}
The contribution analysis of various optimization strategies:
\begin{itemize}
    \item Strategy guidance: approximately 12\% improvement in success rate
    \item Example learning: approximately 8\% improvement in success rate
    \item Output format standardization: 15\% reduction in invalid outputs
\end{itemize}

\subsection{Case Studies}

\textbf{Successful Case:}
\begin{lstlisting}[caption=Successful Decision Case, breaklines=true]
Input State: Player position (150, 380), 
Target platform (280, 380, 360)
Agent Analysis: Target center position 320, horizontal 
distance 170, considering parabolic trajectory, 
recommended force 85
Actual Result: Successful landing, score +1
\end{lstlisting}

\noindent\textbf{Failure Case:}
\begin{lstlisting}[caption=Failure Decision Case, breaklines=true]
Input State: Player position (200, 380), 
Target platform (400, 380, 480)
Agent Analysis: Long distance, recommended force 100
Actual Result: Over-jumping, missed platform
\end{lstlisting}

\section{Limitations and Conclusion}
The limitations of this method include: \textbf{Computational Precision Constraints}: LLMs may exhibit errors in numerical calculations, particularly in complex physical modeling scenarios. \textbf{Real-time Performance Issues}: Each decision requires LLM API calls, introducing latency unsuitable for games requiring extremely high real-time performance. \textbf{Consistency Problems}: LLMs may produce different outputs for identical inputs, affecting decision stability~\cite{chahal2021india,souppaya2024secure,clark2021measuring}.

\noindent\textbf{Conclusion}: the experimental results demonstrate that LLM agents can achieve satisfactory performance in structured game environments through careful prompt engineering. However, challenges remain in computational accuracy, consistency, and real-time performance that require further investigation and improvement.

\bibliographystyle{named}
\bibliography{ijcai25}

\end{document}